\documentclass[journal]{IEEEtran}
\makeatletter\if@twocolumn\PassOptionsToPackage{switch}{lineno}\else\fi\makeatother

\usepackage{graphicx}
\usepackage[T1]{fontenc}
\ifCLASSINFOpdf
\else
\fi
\usepackage{amsmath}
\usepackage[cmintegrals]{newtxmath}
\usepackage{url}

\usepackage{url,multirow,morefloats,floatflt,cancel,tfrupee}
\makeatletter

\AtBeginDocument{\@ifpackageloaded{textcomp}{}{\usepackage{textcomp}}}
\makeatother
\usepackage{colortbl}
\usepackage{xcolor}
\usepackage{pifont}
\usepackage[nointegrals]{wasysym}
\makeatletter
\usepackage{subcaption}
\usepackage{siunitx}

\def\mcWidth#1{\csname TY@F#1\endcsname+\tabcolsep}

\def\cAlignHack{\rightskip\@flushglue\leftskip\@flushglue\parindent\z@\parfillskip\z@skip}
\def\rAlignHack{\rightskip\z@skip\leftskip\@flushglue \parindent\z@\parfillskip\z@skip}

\@ifundefined{etal}{}{}

\usepackage{ifxetex}
\ifxetex\else\if@twocolumn\@ifpackageloaded{stfloats}{}{\usepackage{dblfloatfix}}\fi\fi

\AtBeginDocument{
\expandafter\ifx\csname eqalign\endcsname\relax
\def\eqalign#1{\null\vcenter{\def\\{\cr}\openup\jot\m@th
  \ialign{\strut$\displaystyle{##}$\hfil&$\displaystyle{{}##}$\hfil
      \crcr#1\crcr}}\,}
\fi
}

\AtBeginDocument{%
  \@ifpackageloaded{endfloat}%
   {\renewcommand\efloat@iwrite[1]{\immediate\expandafter\protected@write\csname efloat@post#1\endcsname{}}}{\newif\ifefloat@tables}%
}%

\def\BreakURLText#1{\@tfor\brk@tempa:=#1\do{\brk@tempa\hskip0pt}}
\let\lt=<
\let\gt=>
\def\processVert{\ifmmode|\else\textbar\fi}

\@ifundefined{subparagraph}{
\def\subparagraph{\@startsection{paragraph}{5}{2\parindent}{0ex plus 0.1ex minus 0.1ex}%
{0ex}{\normalfont\small\itshape}}%
}{}

\newcommand\role[1]{\unskip}
\newcommand\aucollab[1]{\unskip}
  
\@ifundefined{tsGraphicsScaleX}{\gdef\tsGraphicsScaleX{1}}{}
\@ifundefined{tsGraphicsScaleY}{\gdef\tsGraphicsScaleY{.9}}{}
\def\checkGraphicsWidth{\ifdim\Gin@nat@width>\linewidth
	\tsGraphicsScaleX\linewidth\else\Gin@nat@width\fi}

\def\checkGraphicsHeight{\ifdim\Gin@nat@height>.9\textheight
	\tsGraphicsScaleY\textheight\else\Gin@nat@height\fi}

\def\fixFloatSize#1{}
\let\ts@includegraphics\includegraphics

\def\inlinegraphic[#1]#2{{\edef\@tempa{#1}\edef\baseline@shift{\ifx\@tempa\@empty0\else#1\fi}\edef\tempZ{\the\numexpr(\numexpr(\baseline@shift*\f@size/100))}\protect\raisebox{\tempZ pt}{\ts@includegraphics{#2}}}}

\AtBeginDocument{\def\includegraphics{\@ifnextchar[{\ts@includegraphics}{\ts@includegraphics[width=\checkGraphicsWidth,height=\checkGraphicsHeight,keepaspectratio]}}}

\DeclareMathAlphabet{\mathpzc}{OT1}{pzc}{m}{it}

\def\URL#1#2{\@ifundefined{href}{#2}{\href{#1}{#2}}}

\def\UrlOrds{\do\*\do\-\do\~\do\'\do\"\do\-}%
\g@addto@macro{\UrlBreaks}{\UrlOrds}

\edef\fntEncoding{\f@encoding}

\makeatother

\newif\ifmultipleabstract\multipleabstractfalse%
\newcommand\notsotiny{\@setfontsize\notsotiny{6}{7}}

\makeatletter
\AtBeginDocument{\@ifpackageloaded{longtable}{%
\def\LT@makecaption#1#2#3{%
  \LT@mcol\LT@cols c{\hbox to\z@{\hss\parbox[t]\LTcapwidth{%
    \sbox\@tempboxa{#1{#2: } #3}%
    \ifdim\wd\@tempboxa>\hsize
      #1{#2: }\textsc{#3}%
    \else
      \hbox to\hsize{\hfil\box\@tempboxa\hfil}%
    \fi
    \endgraf\vskip\baselineskip}%
  \hss}}}
}{}}
\makeatother

   \makeatletter
  \def\fig@textbf{\textbf}
   \AtBeginDocument{\renewcommand\floatc@plain[2]{\setbox\@tempboxa\hbox{{\footnotesize#1.}\footnotesize\hskip.5em#2}%
    \ifdim\wd\@tempboxa>\hsize {\fig@textbf{\footnotesize#1.}}\footnotesize\hskip.5em#2\par
        \else\hbox to\hsize{\hfil\box\@tempboxa\hfil}\fi}}
    \makeatother
  
\usepackage{float}

\usepackage[font=scriptsize]{caption}

\begin{document}

%

        \title{Adversarial Perturbations Prevail in the Y-Channel of the YCbCr Color Space}
      
\author{Camilo~Pestana,
        Naveed~Akhtar,
        Wei~Liu,
        David~Glance, and 
        Ajmal~Mian\thanks{Camilo~Pestana, Naveed~Akhtar, Wei~Liu,  David~Glance, Ajmal~Mian are with The Department of Computer Science, The University of Western Australia, e-mail: camilo.pestanacardeno@research.uwa.edu.au,
        e-mail: naveed.akhtar@uwa.edu.au,
        e-mail: wei.liu@uwa.edu.au, e-mail: david.glance@uwa.edu.au, e-mail: ajmal.mian@uwa.edu.au}}

\maketitle 

\begin{abstract}
Deep learning offers state of the art solutions for image recognition. However, deep models are vulnerable to adversarial perturbations in images that are subtle but significantly change the model's prediction. In a white-box attack, these perturbations are generally learned for deep models that operate on RGB images and, hence, the perturbations are equally distributed in the RGB color space. 
In this paper, we show that the adversarial perturbations prevail in the Y-channel of the YCbCr space. Our finding is motivated from the fact that the human vision and deep models are more responsive 
to shape and texture rather than color. Based on our finding, we propose a defense against adversarial images. Our defense, coined ResUpNet, removes perturbations only from the Y-channel by exploiting ResNet features in an upsampling framework without the need for a bottleneck. At the final stage, the untouched CbCr-channels are combined with the refined Y-channel to restore the clean image. Note that ResUpNet is model agnostic as it does not modify the DNN structure. ResUpNet is trained end-to-end in Pytorch and the results are compared to existing defense techniques in the input transformation category. Our results show that our approach achieves the best balance between defense against adversarial attacks such as FGSM, PGD and DDN and maintaining the original accuracies of VGG-16, ResNet50 and DenseNet121 on clean images. We perform another experiment to show that learning adversarial perturbations only for the Y-channel results in higher fooling rates for the same perturbation magnitude.

\end{abstract}
\vspace{-3mm}
    
%
\IEEEpeerreviewmaketitle

\section{Introduction}

 Convolutional Neural Networks (CNNs) have been effectively used in a wide range of computer vision tasks~\cite{krizhevsky2012ImageNet} including object detection, image classification, and semantic segmentation. However, since 2014~\cite{szegedy2013intriguing} an increasing number of researchers have demonstrated that CNNs are vulnerable to small perturbations in the input image that form  adversarial examples. As more sophisticated attacks are created~\cite{Rony_2019_CVPR}, there is a need for more robust models and better defenses~\cite{erichson2019jumprelu}.
In terms of defenses, adversarial training~\cite{tramer2017ensemble},~\cite{kurakin2016adversarial}, architectured design changes, and input image transformations~\cite{buades2005non} are three approaches that are widely used. In the first approach, adversarial examples are added to the training dataset and the model is re-trained with this new dataset. As a consequence of adversarial training, the model becomes more robust against the used adversarial attack~\cite{tramer2017ensemble}. The second approach is an active area of research usually involving better architecture designs or special activation functions to improve robustness of the model~\cite{xie2019feature}. For the third approach, the image is preprocessed, applying different transformations in an attempt to remove the adversarial patterns in the input~\cite{prakash2018deflecting}. Usually input transformation defenses are agnostic to the targeted model and can be applied to any model as opposed to adversarial training which is specific to individual models.
There are many image transformations that can be used to remove noise in images~\cite{guo2017countering}. However, some of those transformations are not able to remove the adversarial patterns completely. Raff et al.  ~\cite{Raff_2019_CVPR} show that random combination of transformations can help in terms of defense against adaptative attacks. Nevertheless, application of multiple transformations also has significant adverse effects on the model accuracy on clean images.
\vspace{-3mm}
\\
\\
Previous state-of-the-art defenses in the category of image preprocessing usually undertake transformations of colored images in each of the RGB channels of the color space. In this paper, we explore the effects of common adversarial attacks and how the perturbation is allocated in different channels of color spaces such as RGB and YCbCr. The main contributions of our work are listed below:

\begin{figure}[t]
   \vspace{-4mm}
   \centering
   \includegraphics[width=3.5in, height=3.5in]{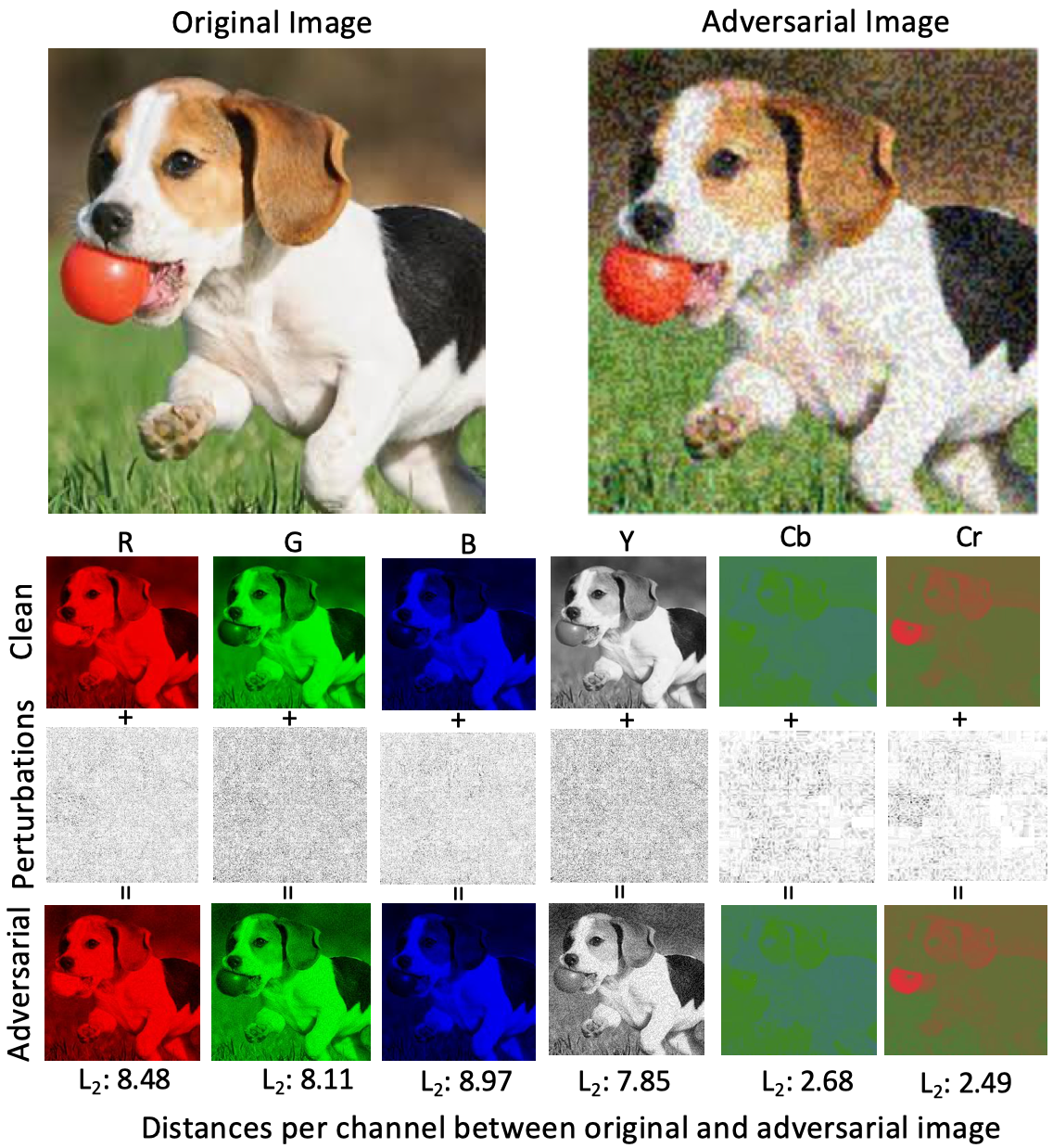} 
   \vspace{-5mm}
   \caption{Comparison between the original and the adversarial images in RGB and YCbCr color spaces. Adversarial images were created with FGSM attack ($\epsilon$ = 0.04).  Per channel $\ell_2$ distances show that Y-channel values are much higher compared to Cb and Cr whereas the R, G and B values are similar.}
\label{fig:ycbcr}
\vspace{-3mm}
\end{figure}

\begin{figure*}[t!]
   \centering
   \includegraphics[width=7.2in, height=3.5in]{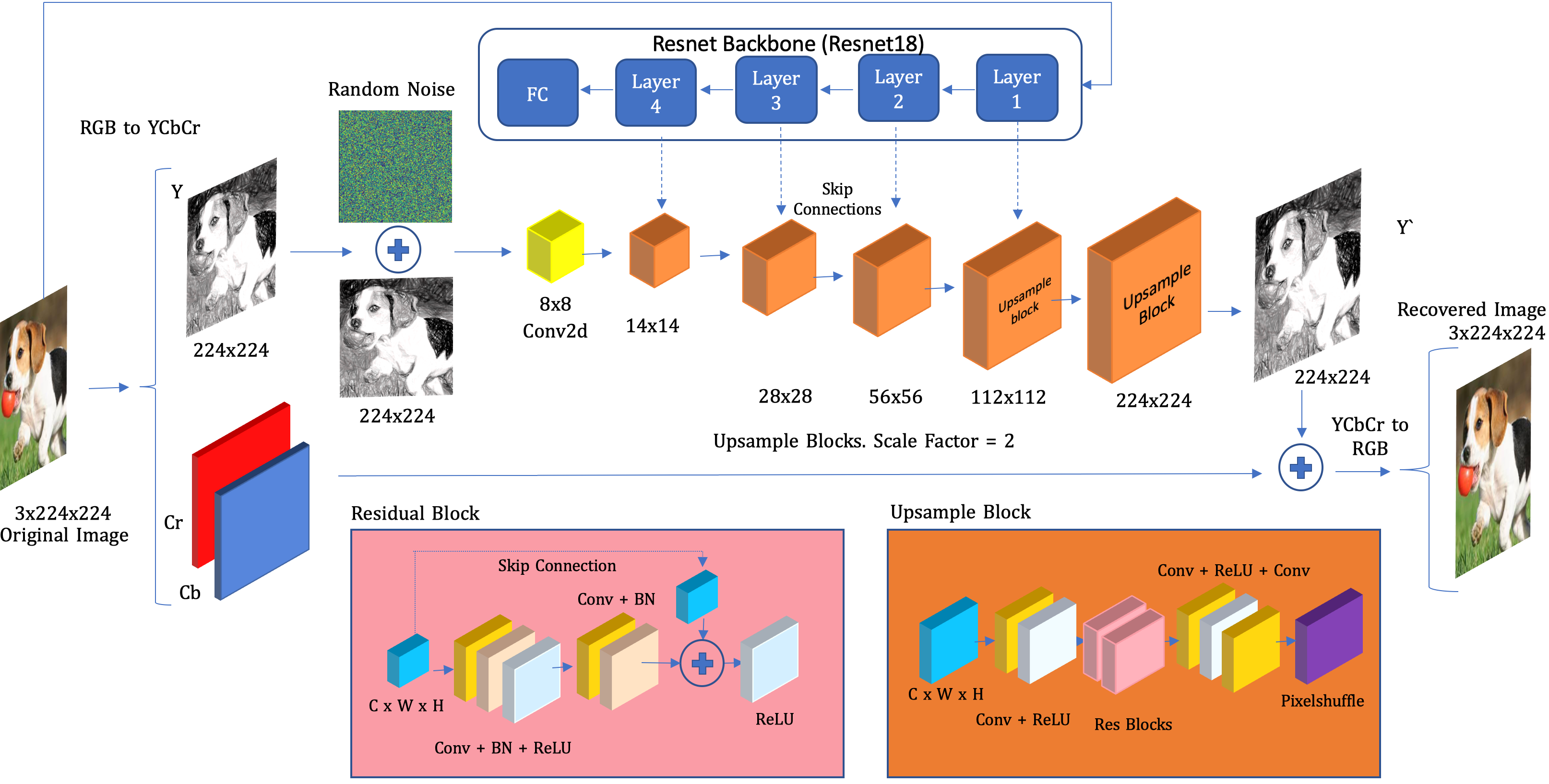} 
   \vspace{-5mm}
   \caption{\textbf{Up-sampling Architecture:} Architecture of Up-sampling model borrowing ResNet features. Those features are passed into the stream of data as a skip connection into the Upsample Blocks. Residual Blocks and Upsample Blocks internal architectures are shown at the bottom of the diagram. In the first step, input images are converted to YCbCr space from an RGB image, Cb and Cr channels remain unchanged until the end of the flow when it is concatenated with the restored Y-channel and converted back to an RGB image. In the case of the Y-channel, random noise is added to it, then passed to a convolutional layer and continues passing through a pipeline of Upsample blocks.}
\label{fig:network}
\vspace{-5mm}
\end{figure*}

1) We show with extensive experiments that adversarial perturbations are allocated more in the Y-channel of the YCbCr color space than in the Cb and Cr channels whereas in the RGB color space, the perturbations are equally distributed between the three channels~(See Figure \ref{fig:ycbcr}).
\\
2) A natural step forward is to develop better defenses that specialize in removing the adversarial perturbation from the Y-channel instead of targeting all channels in the RGB color space. We propose a simple but effective preprocessing defense that borrows features from an auxiliary network such as ResNet to compute their own filters in an Up-sampling Network (Fig. \ref{fig:network}). We pass the image through a convolution layer first, then the up-scaling process starts. Moreover, we only tackle the recovery of the Y-channel from the YCbCr color space. We demonstrate that this approach is able to achieve results comparable to state-of-the-art defenses in the input transformation methods in regards to PGD attacks, while obtaining better results on clean images and other attacks. 
\\
3) Given the importance of the Y-channel for adversarial perturbations, we propose an optimization technique to reduce the amount of perturbation in existing adversarial attacks such as FGSM and PGD while achieving higher fooling rates.
\\

\vspace{-3mm}
\section{Problem Definition}

Let  $x$ $\in$ $\mathbb{R}^{HxWx3}$ denote the original image without perturbations, where H is height, W is width and there are three colour channels, usually (R)ed, (G)reen and (B)lue. Let $y$ $\in$ $\mathbb{R}^n$ be the probabilities of the predicted labels. Given an image classifier $C$ $\rightarrow{\{1,2,...,n\}}$, e.g., for an ImageNet dataset \cite{imagenet_cvpr09} n = 1000, an untargeted attack aims to add a perturbation $p$ to $x$ to compute $x_{adv}$, such that $C(x)\neq C(x_{adv})$, where $p$ could be a grayscale image of the same size $p$ $\in$ $\mathbb{R}^{HxW}$; or a colored image of the same size $p$ $\in$ $\mathbb{R}^{HxWx3}$. The calculation $x_{adv} = x + p$ is constrained such that the perturbation in $x_{adv}$ is imperceptible for the human eye, e.g., $d(x, x_{adv}) \leq \epsilon$ for a distance function $d()$ and a small value $\epsilon$. In the context of adversarial attacks, the distance metric $d()$ is the $L_{p}$ norm of the difference between the original image $x$ and the adversarial image $x_{adv}$. In this paper, we will consider attacks that optimise the $L_{2}$ and $L_{\infty}$ norms only. If the computed $x_{adv}$ satisfies the condition $C(x) \neq C(x_{adv})$ under the given set of constraints, the attack is considered successful. In addition, let D() denote a defense function; if $C(x)\neq C(x_{adv})$ then D() should ideally behave such that $C(D(x_{adv})) = C(x)$.
\vspace{-3mm}

\section{Related Work}

Below, we present review popular adversarial attacks and defenses proposed in the literature, which form the basis of our evaluations and are necessary for understanding our proposed defense mechanism. We only focus on adversarial examples in the domain of image classification, although 
adversarial examples can be created for different tasks such as object recognition ~\cite{li2019rosa} and semantic segmentation ~\cite{xie2017adversarial}. In addition, upsample techniques and input transformation methods are explored in the literature, which are the underlying mechanism of our proposed defense.  

\vspace{-3mm}
\subsection{Attack Algorithms}
Depending on the adversary's knowledge regarding the model under attack, strategies can be categorised as gradient-based attacks (white-box) or gradient-free attacks (black-box). Gradient-based attacks are more powerful and usually less  computationally expensive, hence a defense against them is practically more meaningful. We use gradient-based attacks to train and test our defense strategy.

\textbf{Fast Gradient Signed Method (FGSM)}:  This attack is a single step attack introduced by Goodfellow et al. ~\cite{goodfellow2014explaining} and one of the first adversarial attacks mentioned in the literature. In Eq. \eqref{eq:fgsm}, $x_{adv}$ is the adversarial image and $x_{benign}$ is the original image.  In the cost function $J(\theta, x_{benign}, y))$, $\theta$ represents the network parameters. Moreover, $\epsilon$ is used to scale the noise and is usually a small number. A sign function is applied to the gradients of the loss with respect to the input image to compute the final perturbation.
\vspace{-2mm}
\begin{equation}
\label{eq:fgsm}
    x_{adv} = x_{benign} + \varepsilon * \textrm{sign}(\nabla x_{benign} J(\theta, x_{benign}, y)
\end{equation}
\vspace{-5mm}

\textbf{Basic Iterative Method (BIM)}: This attack is an extension of FGSM, where $\alpha$ is the amount of perturbation added in each iteration. In addition, Kurakin et al.~\cite{kurakin2016adversarial} used a clip function to reduce the impact of large changes on each pixel. Usually the clip function has a range between [0,1] or [0,255] for images.
\vspace{-2mm}
\begin{equation}
\label{eq:bim}
    x^{adv}_{n+1} = \textrm{clip}_x \left\{
    x^{adv}_{n} + \alpha * \textrm{sign}(\nabla x J(x^{adv}_{n}, y))
    \right\}
\end{equation}

\textbf{Iterative Least-Likely Class Method (ILLC)}:
This method is similar to BIM. However, instead of using a true label to optimize the cost function it uses the `least-likely' label. In general, there are untargeted and targeted adversarial attacks. An untargeted attack aims to find an adversarial image that makes a DNN misclassify an image to any incorrect label. On the other hand, targeted attacks misclassify an image to a specific label. ILLC works as a targeted attack using the least-likely label. In order to perform a targeted attack it is necessary to change the sign as shown in Eq. \ref{eq:illc}.
\\
\vspace{-2mm}
\begin{equation}
\label{eq:illc}
    x^{adv}_{n+1} = \textrm{clip}_{x} \left\{
    x^{adv}_{n} - \alpha * \textrm{sign}(\nabla x J(x^{adv}_{n}, y_{LL}))
    \right\}
\end{equation}
\vspace{-5mm}

\textbf{Projected Gradient Descent (PGD):} A stronger iterative version of FGSM is the PGD attack, which is considered one of the strongest attacks and it is used as a benchmark to measure the robustness of many defenses in the literature~\cite{madry2017towards}.

\textbf{Carlini and Wagner (C\&W):}
Carlini and Wagner introduced three attacks in response to one of the first adversarial defenses in the literature called Defensive distillation~\cite{carlini2017towards}. These attacks are hardly noticeable by humans, given that they constrain the $L_{p}$-norm. C\&W attacks are shown to be highly effective against defensive distillation and other types of defenses. C\&W attack that is constrained by $L_{2}$ norm is considered one of the strongest attacks in the literature \cite{rony2019decoupling}, however, this attack can be computationally expensive, often requiring thousands of iterations to calculate an adversarial image.

\textbf{Decoupling Direction and Norm (DDN):}
The DDN attack is an efficient method that was introduced by Rony et al. ~\cite{rony2019decoupling}. DDN optimizes the number of iterations needed while achieving state-of-the-art results, even comparable to C\&W $L_{2}$ constrained attacks.
\vspace{-3mm}
\subsection{defenses}
\vspace{-1mm}
Recently, a number of studies have explored ideas to make DNNs more robust against adversarial attacks. Many of those defense methods are still unable to achieve true robustness to all adversarial inputs. Currently, the most effective defense strategies modify the DNN training process to improve robustness against adversarial examples ~\cite{shaham2018understanding}. However, they are trained to defend against specific attacks, limiting their real-world applications. In contrast, there is another research line which aims to be attack and model agnostic. This line preprocesses the images instead of modifying models or applying specific defenses to them. In general, adversarial defenses can be broken down into three categories:  defenses implemented in the training process ~\cite{stutz2019disentangling},~\cite{taghanaki2019kernelized},~\cite{liu2019rob}, defenses that are applied as an image denoising operation~\cite{raff2019barrage},~\cite{xie2019feature}, and finally architectured design elements or activations that make DNNs more robust, such as JumpReLu ~\cite{erichson2019jumprelu}. Below we describe some of the defenses with respect to these three categories, however, our main focus in this paper are preprocessing methods and these form the underlying concept that we use in our defense.
\vspace{-6mm}
\\
\\
\subsubsection{\textbf{Adversarial Training}}
 The aim of adversarial training is to increase the DNN's robustness by adding adversarial images to the training set~\cite{goodfellow2014explaining},~\cite{lyu2015unified},~\cite{shaham2018understanding}. These methods effectively enhance robustness against adversarial attacks, but they lack generalisation to unknown attacks ~\cite{tramer2017ensemble}. Many different techniques to generate adversarial examples can be applied for data augmentation. Training the model with the original data and the augmented data might increase the DNN's robustness. However, as demonstrated in ~\cite{narodytska2016simple} and ~\cite{papernot2017practical}, this defense is not robust to black-box attacks. 
\\
\subsubsection{\textbf{Image Denoising}}
The approach of pre-processing input examples to remove adversarial perturbations has the advantage of being model-agnostic, and it can be used with any other defense strategies. Bhagoji et al. ~\cite{bhagoji2017dimensionality} proposed a Principal Component Analysis (PCA) method on images to reduce their dimentionality and therefore reduce noise. Alternatively, Das et el. ~\cite{das2017keeping} proposed leveraging JPEG compression as a pre-processing step for adversarial defenses. 
\vspace{-3mm}
\\
\\
\textbf{Pixel Deflection:} Some defenses instead of trying to remove the adversarial patterns on an image aim to corrupt the perturbation by adding noise. Additive noise is the main idea behind some defenses such as the pixel deflection defense ~\cite{prakash2018deflecting}.
\vspace{-3mm}
\\
\\
\textbf{Barrage of Random Transformations (BaRT):}
There have been many attempts to use single input transformations aiming to remove adversarial patterns in images. Some of the methods proposed in the literature include Feature Squeezing ~\cite{xu2017feature}, Blurring filters~\cite{cohen2019certified} and JPEG Compression~\cite{dziugaite2016study},~\cite{guo2017countering}, which are considered weak defenses given that they do not stand against strong attacks with a significant perturbation. He et al. showed that combining weak defenses does not create a stronger defense~\cite{he2017adversarial}. However, Raff et al. demonstrated that randomly selecting transformations from a big pool of transformations into a single barrage (BaRT) is a defense capable of resisting the strongest attacks such as PGD ~\cite{raff2019barrage}. In the BaRT implementation, 25 different transformations are available to randomly select from that pool. The parameters for every transformation are selected randomly as well as the order k of the number of transformations selected. Then, once it is combined in a single pipeline the transformations are applied sequentially.
\\
\\
\vspace{-5mm}
\subsubsection{\textbf{Retrofit defenses}}
Adversarial training requires re-training a DNN, which can be sometimes computationally expensive. Alternatively, there is a line of research which tries to make changes to the architecture of a DNN at inference time by modifying or activating different functions. For example, Erichson et al.~\cite{erichson2019jumprelu} proposed a simple and inexpensive strategy by introducing the new activation function JumpReLU. This activation function uses a ReLU activation and uses a hyperparameter at validation time to control the jump size. They demonstrated empirically that this activation function increases model robustness, protecting against adversarial attacks with high levels of perturbations in MNIST and CIFAR10 datasets.
\vspace{-6.5mm}
\subsection{Up-sampling Networks}
 Up-sampling refers to any technique that converts a low resolution image to a higher resolution image \cite{schulter2015fast}. Up-sampling Networks are used for different computer vision problems such as super-resolution, denoising and image inpainting. For our defense architecture we took inspiration from techniques used in super-resolution, U-Net architectures and Auto-encoders. The concepts below provide the key to understand our defense.
 
\textbf{Super-resolution}:
Super-resolution aims to restore high-resolution images using one or more low-resolution images~\cite{park2003super}. More formally, given a low-resolution image $I_{lr}$, the aim of super-resolution is to generate a high resolution image $I_{hr}$, scaled by a factor number $f \in \mathbb{R}$, such that the result of $I_{hr} = I_{lr} * f $ is an enhanced image with better perceptual quality. In general, $I_{lr}$ is calculated from $I_{hr}$ following a downsampling function $I_{lr} = D(I_{hr}, \delta)$ where $\delta$ represents the parameters for the downsampling function D~\cite{wang2019deep}. Super-resolution has been around for decades~\cite{freeman2002example}, and many approaches have been tried from sparse coding \cite{bhandari2018sampling} to deep learning methods~\cite{yang2019deep}. Having a wide range of applications such as medical imaging ~\cite{greenspan2008super} and surveillance ~\cite{zhang2010super}, super-resolution is an important processing technique in computer vision for enhancing images. Some super-resolution techniques assume there are multiple low-resolution images available from the same scene with different perspectives ~\cite{kawulok2019deep}. In contrast, single image super-resolution (SISR) learns to recover high resolution images from a single low-resolution image. 

\textbf{Pixel-Shuffle}
 Pixel-shuffle~\cite{shi2016real} is described as an efficient sub-pixel convolution layer that learns an array of upsample filters mapping low resolution images into the high resolution output. This allows this layer to learn more complex filters than previous techniques such as bicubic upscale while being an order of magnitude faster than other CNN-based approaches.
 
\textbf{U-Net architecture}
Another type of upscaling operation occurs in Autoencoders~\cite{folz2018adversarial} or architectures such as UNet \cite{ronneberger2015u}. In these architectures, an image is downscaled and encoded in a vector or bottleneck. From that bottleneck the next part of the architecture scales up and passes the features through learnable filters. UNet architectures differ from traditional Autoencoders by using skip connections from the downsampling side across the upscaling network.
\vspace{-3mm}
\\
\\
\section{Perturbations in Y-Channel}
Adversarial perturbations are usually computed in the RGB color space. However, there are benefits to using different color spaces in regards to adversarial attacks and defenses in DNNs. For example, YCbCr conveniently separates the
luma channel Y from chroma channels Cb and Cr. The JPEG compression algorithm uses YCbCr color space in its compression process
reducing the color of an image in
a process called Chroma Subsampling \cite{poynton2002chroma} while retaining information from the Y-channel, which is
considered more ‘relevant’ to human perception. This comes from the fact that the Y-channel provides more shape and texture related information than the other two channels which are focused on color (See Fig. \ref{fig:ycbcr}).  Recent research also suggests  that  ImageNet-trained  CNNs  are biased towards texture \cite{geirhos_imagenet-trained_2019}, and contrary to some common beliefs, texture  is  more  important  for object recognition classifiers than other image elements such as color. Moreover, adversarial attacks are optimization problems that usually constrain the amount of perturbation to make adversarial images imperceptible for the human eye. These attacks try to find the most efficient way to allocate perturbations in an image with the least amount of perturbation. If shape and texture features are more relevant for ImageNet-based CNNs, it also means that perturbing these features should be more efficient. Therefore, if we use YCbCr color space, more  perturbation  should  be  concentrated  in  the Y-channel  rather  than  in  the  color  channels  Cb  and  Cr. On the contrary, given that the shape and texture features are spread equally in the RGB color space, there should be a similar amount of perturbation in each channel for this color space. 

To validate the above claims, we perform an experiment on 5000 randomly selected images from the ImageNet validation dataset 
and generate adversarial images using attacks FGSM, PGD and DDN with different levels of perturbations ranging between $\epsilon=0.01$ and $\epsilon=0.04$. Next, we calculate the per channel  $\ell_2$ difference between the original image and the adversarial image in the YCbCr and RGB color spaces.

There are several standards for YCbCr conversion. We focus on the standard specified in JFIF version 1.02 \cite{hamilton2004jpeg} and according to it, channels Y, $C_{b}$, and $C_{r}$ have a full range [0,255]. Equations \ref{eq:ycbcr} and \ref{eq:rgb} show the relationship between RGB and YCbCr.
\\
\\
\vspace{-5mm}
\begin{equation}
\label{eq:ycbcr}
\small
    \begin{array}{l}
    Y = 0 + (0.299 R) + (0.587 G) + (0.114 B) \\
    C_{b} = 128 - (0.168736 R) - (0.331264 G) + (0.5 B) \\
    C_{r} = 128 + (0.5 R) - (0.418688 G) - (0.081312 B)
    \end{array}
    \vspace{-2mm}
\end{equation}

\begin{equation}
\label{eq:rgb}
\small
    \begin{array}{l}
    R = Y + 1.402 (C_{r} - 128) \\
    G = Y - 0.344136 (C_{b} - 128) - 0.714136 (C_{r} - 128) \\
    B = Y + 1.772 (C_{b} - 128)
    \end{array}
\end{equation}

\begin{table*}[htbp]
  \centering
  \setlength\tabcolsep{1.5pt}
  \caption{\textbf{Color space perturbations}:
  This table shows the results for a subset of 5000 images from the ImageNet validation dataset (Dataset-2. See Table \ref{tab:accurnodefense}). In this table, we take the average of the difference between original images and their adversarial versions under adversarial attacks FGSM, PGD and DDN with epsilon values between $\epsilon$=0.01~and~$\epsilon$=0.04. }
    \begin{tabular}{|c|c|c|c|c|c|c|c|c|c|c|c|c|c|c|c|c|c|c|c|c|c|c|c|c|c|c|c|}
    \hline
    \rowcolor[rgb]{ .851,  .851,  .851}       & \multicolumn{9}{c|}{FGSM}                                             & \multicolumn{9}{c|}{PGD}                                              & \multicolumn{9}{c|}{DDN} \\
    
    \rowcolor[rgb]{ .851,  .851,  .851} MODELS & \multicolumn{3}{c|}{e=0.01} & \multicolumn{3}{c|}{e=0.02} & \multicolumn{3}{c|}{e=0.04} & \multicolumn{3}{c|}{e=0.01} & \multicolumn{3}{c|}{e=0.02} & \multicolumn{3}{c|}{e=0.04} & \multicolumn{3}{c|}{n=20} & \multicolumn{3}{c|}{n=40} & \multicolumn{3}{c|}{n=60} \\
    
    \rowcolor[rgb]{ .851,  .851,  .851} YCbCr & Y     & Cb    & Cr    & Y     & Cb    & Cr    & Y     & Cb    & Cr    & Y     & Cb    & Cr    & Y     & Cb    & Cr    & Y     & Cb    & Cr    & Y     & Cb    & Cr    & Y     & Cb    & Cr    & Y     & Cb    & Cr \\
    
    Vgg16 & 6.30  & 4.19  & 4.15  & 7.48  & 4.76  & 4.68  & 8.51  & 5.22  & 5.11  & 6.60  & 4.40  & 4.46  & 7.95  & 5.13  & 5.19  & 9.26  & 5.80  & 5.94  & 2.89  & 1.92  & 2.09  & 2.72  & 1.83  & 1.00  & 2.51  & 1.72  & 1.89 \\
    
    Resnet50 & 6.40  & 4.24  & 4.20  & 7.58  & 4.78  & 4.70  & 8.60  & 5.20  & 5.10  & 6.50  & 4.40  & 4.52  & 7.86  & 5.16  & 5.21  & 9.09  & 5.98  & 6.10  & 2.89  & 1.92  & 2.09  & 2.72  & 1.83  & 2.00  & 2.51  & 1.72  & 1.89 \\
    
    Densenet121 & 6.40  & 4.27  & 4.30  & 7.13  & 4.85  & 4.88  & 8.10  & 5.33  & 5.35  & 6.30  & 4.40  & 4.52  & 7.73  & 5.19  & 5.28  & 9.02  & 6.01  & 6.16  & 2.89  & 1.92  & 2.09  & 2.72  & 1.83  & 2.10  & 2.18  & 1.54  & 1.74 \\
    
    \rowcolor[rgb]{ .851,  .851,  .851} RGB   & R     & G     & B     & R     & G     & B     & R     & G     & B     & R     & G     & B     & R     & G     & B     & R     & G     & B     & R     & G     & B     & R     & G     & B     & R     & G     & B \\
    
    Vgg16 & 0.26  & 0.26  & 0.26  & 0.51  & 0.51  & 0.51  & 1.02  & 1.02  & 1.02  & 0.22  & 0.20  & 0.20  & 0.38  & 0.38  & 0.37  & 0.69  & 0.69  & 0.69  & 0.03  & 0.04  & 0.05  & 0.03  & 0.03  & 0.04  & 0.03  & 0.03  & 0.04 \\
    
    Resnet50 & 0.26  & 0.26  & 0.26  & 0.51  & 0.51  & 0.51  & 1.01  & 1.01  & 1.02  & 0.20  & 0.20  & 0.20  & 0.38  & 0.38  & 0.37  & 0.69  & 0.69  & 0.69  & 0.03  & 0.04  & 0.05  & 0.03  & 0.03  & 0.04  & 0.03  & 0.03  & 0.04 \\
    
    Densenet121 & 0.26  & 0.26  & 0.26  & 0.51  & 0.51  & 0.51  & 1.01  & 1.01  & 1.02  & 0.20  & 0.20  & 0.20  & 0.38  & 0.38  & 0.37  & 0.69  & 0.69  & 0.69  & 0.03  & 0.04  & 0.05  & 0.03  & 0.03  & 0.04  & 0.03  & 0.03  & 0.04 \\
    \hline
    \end{tabular}%
  \label{tab:colorspaces}%
\end{table*}%

Our results in Table \ref{tab:colorspaces} corroborates the intuition presented above and show that images on the RGB space have a very similar amount of perturbation in each channel. However, in the YCbCr color space, the Y-channel consistently presents more perturbation than the Cb and Cr channels. We also illustrate this fact qualitatively in Figure \ref{fig:ycbcr}. Whereas Table~\ref{tab:colorspaces} shows the average values, we confirmed that the Y-channel distance was significantly higher than the Cb and Cr channels for each one of the 5000 images.

Since perturbations are allocated more strongly in the Y-channel and this channel contains more relevant features, we propose a method for removing perturbations only from the Y-channel in an effort to make existing CNN models robust to adversarial attacks while maintaining their accuracies on clean images.

\section{Proposed Approach}

The aim of our proposed approach is achieve model-agnostic defense by preprocessing an input image and then reconstructing it with an Upsampling Network while removing adversarial patterns. Our defense consists of a single transformation that uses a ResNet model~\cite{he2016deep} pretrained on ImageNet dataset to extract features from its internal layers. Our architecture takes inspiration from U-Net~\cite{ronneberger2015u} and Dynamic U-Nets proposed by the FastAI research team~\cite{howard2018fastai}. Similar to U-Net architecture,  skip connections are used to concatenate the features from the encoder layers to the upsampling network. However, it differs from traditional autoencoders by not using a bottleneck, which is essential for these architectures. Instead of training a decoder from scratch, a pretrained ResNet18 model is used to extract meaningful features and add them to the main Upsample Network. Given that meaningful features are passed directly from the auxiliary network to our DNN at different stages, the use of a bottleneck becomes redundant.

ResNet18 is the shallowest version from the original proposed ResNet architectures. However, we can extract useful feature maps from their internal layers. While using deeper architectures such as ResNet-101 or ResNet-152 might be beneficial, providing more interesting and robust features, it will also increase the number of parameters and computational time to process each image, which is not desirable. Therefore, we demonstrate that using a shallow ResNet backbone is enough to create a strong defense. In addition, as part of the preparation to train the network images, our network adds random noise to every input image in the training and evaluation step. As mentioned in Section III-B, some defenses have used additive random noise as part of their approach, and recent research suggests that additive random noise can significantly improve robustness of a model~\cite{li2019certified}. Moreover, the Upsampling Network selectively creates high frequency components and recovers the components, removing at the same time adversarial perturbations. Once the image is preprocessed using the Upsampling model, the new image version can be passed to different models. 
\begin{figure*}[htbp]
   \centering
   \includegraphics[width=7.2in, height=3.5in]{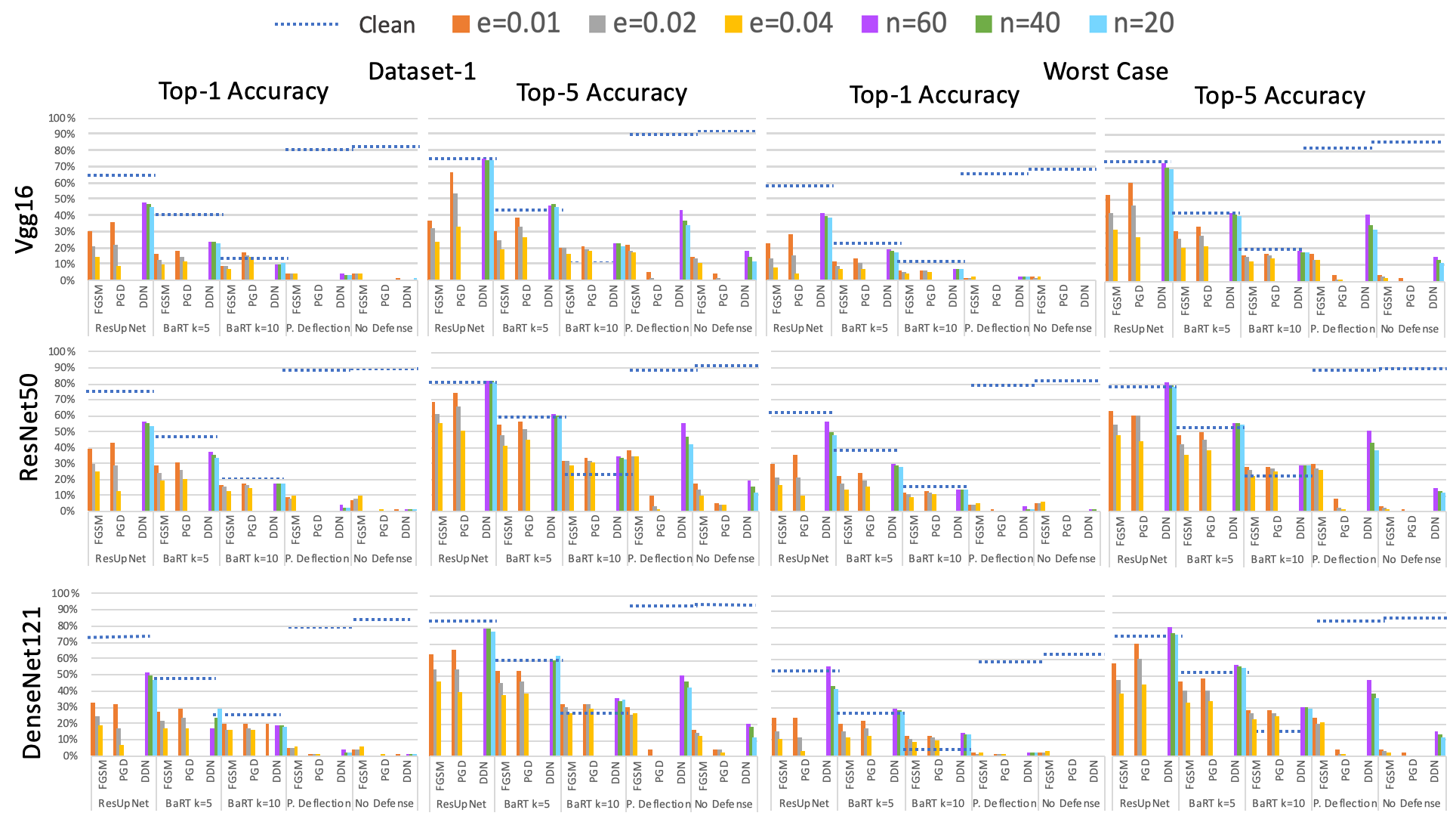} 
   \caption{ The figure illustrates defenses such as Pixel Deflection, BaRT k=5, BaRT k=10 and \textbf{ResUpNet (ours)} under FGSM, PGD and DDN attacks. These defenses are evaluated using Dataset-1 and Worst in accordance with Table \ref{tab:accurnodefense}. Accuracy for each defense is measure under different values of perturbation $\epsilon=0.01$, $\epsilon=0.02$, $\epsilon=0.04$. Given that DDN attack works differently to generate adversarial images, instead of using $\epsilon$, a number of iterations $n$ with parameters $n=60$, $n=40$ and $n=20$ is used. For both datasets under evaluation, Top-1 and Top-5 accuracy are evaluated for models Vgg16, ResNet50 and DenseNet121. }
\label{fig:results}
\vspace{-6mm}
\end{figure*}
\vspace{-3mm}
\subsection{Implementation Details}
We implemented our defense architecture using the Python library Pytorch~\cite{paszke2017automatic}. Moreover, we used Advertorch~\cite{ding2019advertorch} to create adversarial images for three different architectures Vgg16~\cite{simonyan2014very}, ResNet50~\cite{he2016deep}, and DenseNet121~\cite{huang2017densely}. Those models have unique architecture mechanisms and differ in the number of layers, which makes them a good benchmark to test our defense with contrasting models. For $L_\infty$ constrained attacks such as FGSM and PGD, we use $\epsilon$ values of 0.01, 0.02 and 0.04 for images with values in the range [0,1]. In the case of DDN attack, instead of $\epsilon$, we use the number of iterations $n$ as the hyper-parameter, with values 20, 40, and 60.
\vspace{-2mm}
\\
\\
For the training process, given that we are borrowing features from a pretrained model such as ResNet18, we only used the ImageNet validation dataset that contains 50000 images and divided it into two subsets. We used 40000 images for training our defense and 10000 images for validation. In the training process, before processing each batch, we randomly selected between FGSM and PGD attack as well as random perturbations ranging from $\epsilon$ = 0 to $\epsilon$ = 0.05 and applied the attacks to each batch. In addition, we used Mean Square Error loss, and Adam as an optimizer with learning rate 0.0001. We trained our defense for 15 epochs. Each epoch took just over 5 hours in total using a GeForce RTX 2080 ti GPU. The code for our defense and other useful scripts will be available at http://www.github.com/elcronos/ResUpNet.

\vspace{-3mm}
\subsection{ResUpNet Architecture}

Figure \ref{fig:network} illustrates the architecture used for the defense mechanism. In this architecture, we use Residual Blocks that contain Convolution layers, Batch Normalization and ReLU activations. These Residual Blocks are then used in the Upsample Blocks which introduces a PixelShuffle layer at the end of the block for the upsampling operation.
\vspace{-3mm}
\\
\\
Our approach follows two streams. In the first stream, the input image in the RGB space is processed by the ResNet Backbone. In this step, we extract the features from four different layers and save them in memory. In the second stream, the input image is converted from RGB space to YCbCr space. The Y-channel is processed while Cb and Cr channels remain unchanged. Continuing with the process, random noise is added to the Y-channel signal and then passed through a Convolutional layer. Many defenses have used noise injection as part of their defense strategy and recent research suggests that it can improve DDNs robustness against adversarial attacks for white-box and black-box attacks~\cite{araujo2019robust}~\cite{rakin2018parametric}. Even though our defense focus is on white-box attacks, some researchers have demonstrated the effectiveness of using randomization in defenses, which can be specially effective against black-box attacks hardening the process of estimating the gradients of our defense ~\cite{potdevin2019empirical},~\cite{gu5068towards},~ \cite{Raff_2019_CVPR}. After this, Upsample Blocks are used with a \textit{scale factor = 2} to scale up the Y-channel. In the process, the data extracted from the ResNet Backbone is shared with those Upsample Blocks using skip connections. Finally, the resulting recovered Y-channel is concatenated with the unchanged Cb and Cr channels and the image is converted back to RGB.

\section{Experiments}
We perform two types of experiments. In the first one, we show results of our ResUpNet. In the second experiment, we show that by focusing perturbations in the Y-channel, we can achieve higher fooling rates for the same attack types and with the same value of perturbations.

\subsection{Removing Perturbations in the Y-channel}

We tested our defense considering $L_{2}$ and $L_{\infty}$ constrained attacks. The first attack is the well-known one-step attack FGSM. To compare the effect of one-step attacks and iterative attacks, we tested the Projected Gradient Descent (PGD) attack as well, which is considered a strong attack. Additionally, from the $L_{2}$ constrained norm attacks, we used DDN instead of the C\&W attack. DDN provides state-of-the-art attacks while converging much faster and with fewer iterations than the C\&W attack. For the implementation of these attacks, we used the Advertorch library \cite{ding2019advertorch} which integrates seemlessly with Pytorch.

\begin{table}[h!]
  \centering
  \setlength\tabcolsep{2pt} 
  \scriptsize
  \caption{Accuracy of Datasets with no defense:
  Datasets D-(1-5) get very similar results with a very small standard deviation between them. However, dataset `Best' consistently gets better results than the other 5 randomly selected datasets while dataset `Worst' has the lowest performance. }
    \begin{tabular}{|c|l|r|r|r|r|r|r|r|r|r|r|}
    \hline
    \rowcolor[rgb]{ .749,  .749,  .749}       & MODEL & \multicolumn{1}{l|}{CLEAN} & \multicolumn{3}{c|}{FGSM} & \multicolumn{3}{c|}{PGD} & \multicolumn{3}{c|}{DDN} \\
    
    \rowcolor[rgb]{ .749,  .749,  .749} DATASETS &       &       & \multicolumn{1}{c|}{0.01} & \multicolumn{1}{c|}{0.02} & \multicolumn{1}{c|}{0.04} & \multicolumn{1}{c|}{0.01} & \multicolumn{1}{c|}{0.02} & \multicolumn{1}{c|}{0.04} & \multicolumn{1}{c|}{20} & \multicolumn{1}{c|}{40} & \multicolumn{1}{c|}{60} \\
    
    \multirow{3}[6]{*}{Dataset 1} & vgg16 & 71.8  & 4.1   & 4.2   & 4.6   & 0.0   & 0.0   & 0.2   & 1.4   & 0.9   & 0.8 \\
          & resnet50 & 75.9  & 7.4   & 7.6   & 9.6   & 0.0   & 0.0   & 0.0   & 0.4   & 0.2   & 0.1 \\
          & densenet 121 & 74.6  & 4.2   & 4.3   & 5.6   & 0.4   & 0.2   & 0.0   & 0.3   & 0.1   & 0.1 \\
    
    \multirow{3}[6]{*}{Dataset 2} & vgg16 & 70.0  & 4.0   & 3.7   & 4.5   & 0.5   & 0.3   & 0.2   & 1.3   & 0.1   & 0.1 \\
          & resnet50 & 74.8  & 6.7   & 6.3   & 8.0   & 0.0   & 0.0   & 0.0   & 0.3   & 0.1   & 0.1 \\
          & densenet 121 & 73.0  & 3.4   & 3.4   & 4.9   & 0.1   & 0.0   & 0.0   & 0.2   & 0.8   & 0.1 \\
    
    \multirow{3}[6]{*}{Dataset 3} & vgg16 & 71.8  & 4.7   & 4.3   & 4.8   & 0.7   & 0.5   & 0.2   & 1.4   & 1.1   & 1.0 \\
          & resnet50 & 76.3  & 7.8   & 7.6   & 9.4   & 0.1   & 0.0   & 0.0   & 0.6   & 0.3   & 0.2 \\
          & densenet 121 & 74.6  & 4.0   & 3.7   & 5.4   & 0.1   & 0.0   & 0.0   & 0.3   & 0.2   & 0.1 \\
    
    \multirow{3}[6]{*}{Dataset 4} & vgg16 & 71.0  & 4.5   & 4.4   & 5.2   & 1.0   & 0.7   & 0.4   & 1.6   & 1.3   & 1.2 \\
          & resnet50 & 76.3  & 8.0   & 8.0   & 9.8   & 0.1   & 0.0   & 0.0   & 0.4   & 0.2   & 0.1 \\
          & densenet 121 & 74.2  & 4.1   & 4.0   & 5.4   & 0.1   & 0.0   & 0.0   & 0.2   & 0.1   & 0.1 \\
    
    \multirow{3}[6]{*}{Dataset 5} & vgg16 & 71.6  & 4.5   & 4.4   & 4.8   & 0.7   & 0.4   & 0.3   & 1.5   & 1.2   & 1.0 \\
          & resnet50 & 77.0  & 7.0   & 7.5   & 9.2   & 0.0   & 0.0   & 0.0   & 0.3   & 0.1   & 0.0 \\
          & densenet 121 & 74.7  & 4.1   & 3.8   & 5.3   & 0.0   & 0.0   & 0.0   & 0.2   & 0.1   & 0.1 \\
    
    \multirow{3}[6]{*}{Best} & vgg16 & 94.9  & 11.2  & 10.7  & 11.8  & 1.6   & 1.2   & 0.7   & 3.5   & 2.5   & 2.1 \\
          & resnet50 & 96.9  & 18.4  & 17.7  & 22.3  & 0.1   & 0.0   & 0.0   & 1.2   & 0.5   & 0.2 \\
          & densenet 121 & 96.4  & 9.9   & 9.5   & 13.2  & 0.2   & 0.1   & 0.0   & 0.6   & 0.3   & 0.2 \\
    
    \multirow{3}[6]{*}{Worst} & vgg16 & 67.5  & 2.1   & 2.0   & 2.5   & 0.1   & 0.0   & 0.0   & 0.3   & 0.2   & 0.1 \\
          & resnet50 & 73.7  & 4.8   & 4.9   & 5.9   & 0.0   & 0.0   & 0.0   & 0.2   & 0.2   & 0.0 \\
          & densenet 121 & 70.9  & 2.0   & 1.9   & 2.7   & 0.0   & 0.0   & 0.0   & 0.0   & 0.0   & 0.0 \\
    \hline
    \end{tabular}%
  \label{tab:accurnodefense}%
\end{table}%

Initially, we selected 5 subsets of 5000 mutually exclusive images from the ImageNet validation dataset (See Table \ref{tab:accurnodefense}). Akhtar et al.~\cite{akhtar2018defense} argue that evaluating defense mechanisms on already misclassified images is not meaningful and those images should not be considered for evaluation since an attack on a misclassified image is considered successful by default and this could mislead the interpretation of the results. For that reason, some researchers have used a subset on ImageNet that guarantees a 100\% success rate, or greater than 80\% on clean images. We observed in Table \ref{tab:accurnodefense} that the 5 randomly selected datasets have accuracies on clean images around 70\% with a standard deviation close to 1\%. However, we wanted to identify a subset that has greater than 80\% accuracy on clean images. We found that inside the 50000 images in the ImageNet validation dataset, there is a subset of images that performs considerably better on clean images, specially under the FGSM attack, and another dataset that performs worse than the 5 randomly selected sets.
\vspace{-2mm}
\\
\\
In the case of the best scenario, the accuracy for FGSM attack without any defense is around 20\% higher than the rest of the datasets. We also observed that the reported accuracy under an FGSM attack is significantly higher as well, indicating that this particular subset of images seems to be more robust under weak attacks (excluding PGD and DDN). Using the best dataset, which has better initial accuracies on clean images without defenses, could be misleading and it is not a fair representation of the ImageNet dataset. For this reason, we decided to arbitrarily select any of the 5 datasets which have a very low standard deviation between them. We selected Dataset 1. All the indices from each subset of images we created, including the Best and Worst scenarios are in our Github Repository. This facilitates future research to make comparisons with the same datasets.
\vspace{-2mm}
\\
\\
Figure \ref{fig:results} shows quantitative results comparing our defense \textbf{ResUpNet} with other input transformation defenses, including the state-of-the-art defense BaRT with parameters k=5 transformations and k=10 transformations. We used `Dataset-1' and `Worst' (See Table \ref{tab:accurnodefense}) to evaluate those defenses against FGSM, PGD and DDN attack using different levels of perturbations. While Pixel Deflection defense seems to outperform the accuracy compared to other defenses on clean images, it performs poorly against adversarial attacks. Our results also show that ResUpNet performs better or as good as the two BaRT defenses in many cases: all the variations of n's for DDN attack and the other attacks with $\epsilon=0.01$ and $\epsilon=0.02$ are consistently better than the other defenses for `Dataset-1' and `Worst', both in Top-1 and Top-5 accuracy metrics. However, results are not consistent on the $\epsilon=0.04$ cases. Despite this, our defense has the best trade-off between accuracy on clean images and recovery of accuracy under adversarial attacks. 

\subsection{Perturbing only the Y-Channel}
\begin{table}[htbp]
  \centering
  \setlength\tabcolsep{2.8pt}
  \caption{\textbf{Y Attack:} Attacks FGSM, PGD and our versions FGSM-Y and PGD-Y are evaluated using small values for $\epsilon$. The values on the table show the percentage of success attack for different models. For this experiment, we used `Dataset-5'~(See Table \ref{tab:accurnodefense})}
    \begin{tabular}{|c|c|c|c|c|c|c|}
    \hline
    \rowcolor[rgb]{ .851,  .851,  .851} Attack & Models & $\epsilon=$0.002 & $\epsilon=$0.003 & $\epsilon=$0.004 & $\epsilon=$0.005 & $\epsilon=$0.01 \\
    
    \hline
    \multirow{3}[1]{*}{FGSM} & Vgg   & 77.5  & 85.8  & 89.4  & 93.7  & 93.7 \\
          & ResNet & 70.5  & 79.6  & 84.6  & 86.8  & 90.9 \\
          & DenseNet & 78.0  & 84.8  & 89.3  & 91.2  & 94.5 \\
    
    \hline
    \multirow{3}[1]{*}{FGSM-Y} & Vgg   & 88.3  & 96.5  & 98.1  & 98.7  & 100.0 \\
          & ResNet & 81.6  & 94.2  & 97.9  & 99.1  & 100.0 \\
          & DenseNet & 83.4  & 95.4  & 99.0  & 99.5  & 100.0 \\
    
    \hline
    \multirow{3}[1]{*}{PGD} & Vgg   & 69.5  & 80.2  & 85.8  & 89.0  & 93.5 \\
          & ResNet & 63.0  & 73.8  & 80.1  & 84.1  & 90.6 \\
          & DenseNet & 68.7  & 78.8  & 85.0  & 88.4  & 94.1 \\
    
    \hline
    \multirow{3}[1]{*}{PGD-Y} & Vgg   & 86.0  & 95.9  & 98.0  & 98.5  & 100.0 \\
          & ResNet & 78.6  & 93.6  & 97.7  & 99.0  & 100.0 \\
          & DenseNet & 80.4  & 94.4  & 98.5  & 99.5  & 100.0 \\
    \hline
    \end{tabular}%
  \label{tab:yattack}%
\end{table}%

Standard attacks such as FGSM and PGD allocates  more  perturbation in the  Y  channel. To optimize further those attacks, it is possible to create an iterative process selecting only perturbations in the Y-channel and discarding the perturbation in Cb and Cr, hence, we can reduce the amount of perturbation needed for a successful attack. For this experiment, we used FGSM and PGD attacks selecting only the perturbations on the Y-channel.  In  each  iteration of our Y Attack,  we  use a small constant $\alpha$=0.0001 and calculate the perturbation. Then, the adversarial  image,  usually  in  RGB  color  space,  is  converted to  YCbCr  using  Eq.  4.  The  original  image  is  also  converted to  YCbCr  and its Y  channel values are replaced by  the  adversarial Y  channel. The partially perturbed image is then converted back to  RGB  and  the  process  is  repeated until obtaining an adversarial image constrained by $\alpha$, a maximum perturbation $\epsilon$ and maximum number of iterations.  Table \ref{tab:yattack}  show  the comparison between our Y Attack optimization for FGSM and PGD and the original attacks. We call FGSM-Y and PGD-Y the corresponding version of the attack with our Y Attack optimization. For both attacks FGSMI-Y and PGD-Y, we achieve better success rates with less perturbations than the original attacks.

\vspace{-3mm}
\section{Conclusion}
 We introduced a ResUpNet defense, which achieves state-of-the-art results in the category of input transformation defenses. We demonstrate that little training is needed to reconstruct images from adversarial images when we borrow features from pretrained ResNet backbones. Additionally, we demonstrated for the first time that adversarial images on average show much more perturbation concentration in the Y-channel of YCbCr color space and that denoising the Y-channel alone is sufficient to defend from strong attacks such as PGD and DDN. Previous input transformation based defenses in the literature tried to remove adversarial perturbations from the 3 channels in the RGB space. To the best of our knowledge, this is the first adversarial defense that focuses on the removal of perturbations in a single channel and is able to achieve state-of-the-art results while retaining a good ratio between accuracy on clean images and robustness against adversarial images. Our defense being model-agnostic is able to generalize well across different models with different architectures and layers and can potentially be combined with other defenses to improve its effectiveness.  In addition, existent attacks could benefit from this insight as well and try to focus on the Y perturbations to reduce the amount of perturbations needed for a successful attack.
 \vspace{-3mm}
 \section{Acknowledgements}
 The main author was recipient of an Australian Government Research Training Program (RTP) Scholarship at The University of Western Australia. This research is supported by the Australian Research Council (ARC) grant DP190102443.



%

\bibliographystyle{IEEEtran}
\bibliography{references}

\end{document}